\documentclass[conference]{IEEEtran}
\IEEEoverridecommandlockouts

\usepackage{cite}
\usepackage{amsmath,amssymb,amsfonts}
\usepackage{algorithmic}
\usepackage{graphicx}
\usepackage{textcomp}
\usepackage{xcolor}
\usepackage{hyperref}
\usepackage{color}
\usepackage{tabularx}
\usepackage{booktabs}
\usepackage{multirow}
\usepackage{float}
\usepackage{makecell}
\usepackage{caption}
\usepackage{subcaption}
\usepackage{comment}

\makeatletter
\renewcommand\IEEEkeywordsname{Keywords}
\makeatother
\usepackage{xpatch}
\xpatchcmd\IEEEkeywords{---}{-}{}{}

\makeatletter
\renewcommand{\fnum@figure}{Figure~\thefigure}
\makeatother

\def\BibTeX{{\rm B\kern-.05em{\sc i\kern-.025em b}\kern-.08em T\kern-.1667em\lower.7ex\hbox{E}\kern-.125emX}}

\begin{document}

\title{\bfseries\Large Scribe Verification in Chinese manuscripts using Siamese, Triplet, and Vision Transformer Neural Networks}
\author{\IEEEauthorblockN{Dimitrios-Chrysovalantis Liakopoulos\IEEEauthorrefmark{1},Yanbo Zhang\IEEEauthorrefmark{2}, Chongsheng Zhang\IEEEauthorrefmark{2} 
and Constantine Kotropoulos\IEEEauthorrefmark{1}}
\IEEEauthorblockA{\IEEEauthorrefmark{1} 
\textit{Department of Informatics, Aristotle University of Thessaloniki}\\
Thessaloniki 54124, Greece\\
email: \{dliako, costas\}@csd.auth.gr}
\IEEEauthorblockA{\IEEEauthorrefmark{2} 
\textit{School of Computer and Information Engineering, Henan University}\\
Kaifeng, China\\
email: \{zhangyanbo, cszhang\}@henu.edu.cn}}

\maketitle

\begin{abstract}
The paper examines deep learning models for scribe verification in Chinese manuscripts. That is, to automatically determine whether two manuscript fragments were written by the same scribe using deep metric learning methods. 
Two datasets were used: the Tsinghua Bamboo Slips Dataset and a selected subset of the Multi-Attribute Chinese Calligraphy Dataset,  focusing on the calligraphers with a large number of samples. 
Siamese and Triplet neural network architectures are implemented, including convolutional and Transformer-based models. 
The experimental results show that the MobileNetV3+ Custom Siamese model trained with contrastive loss achieves either the best or the second-best overall accuracy and area under the Receiver Operating Characteristic Curve on both datasets. 
\end{abstract}


\section{Introduction}
\label{sec:intro}
Ancient script images exhibit unique characteristics that pose significant challenges to conventional recognition methods. These challenges stem from limited sample availability and image noise arising from the historical and environmental conditions under which ancient scripts were created and preserved.
Many deep learning- and computer vision-based methods for ancient script recognition have emerged, aiming to perform classification, recognition, and interpretation at larger scales and achieve higher accuracy~\cite{Diao2025}.

A modified Convolutional Neural Network (CNN) with an attention mechanism for historical Chinese character recognition is described in \cite{Qin2017}. A batch normalization layer is incorporated into a modified GoogLeNet architecture, and a feature-based attention mechanism is integrated into the network.  Feature vectors are normalized to unit $\ell_2$ norm before the softmax layer, thereby improving generalization.

A CNN is also used to encode a text line image. At the same time, a Bidirectional Long Short-Term Memory network followed by a fully connected neural network serves as the decoder for predicting a sequence of characters in ancient Greek manuscripts \cite{Markou2021}. 
A transformer-based character recognition model incorporating multi-scale attention and Multilayer Perceptron modules is tested on Tsinghua bamboo slips~\cite{Wu2021}. The motivation was to replace the traditional convolution to improve recognition accuracy.

A comparative analysis between various structural and statistical features and classifiers against deep neural networks is performed in \cite{Sivan2025}.  The incorporation of the Discrete Cosine Transform with the Vision Transformer (ViT) is shown to achieve the top $F_1$ score.  

Scribe verification focuses on determining whether two manuscript fragments were written by the same individual. 
In the field of digital humanities, this task plays a significant role in the preservation, classification, and authentication of ancient texts, enabling researchers to group manuscript fragments by writing style and origin.

The main objective of the paper is to develop a system capable of automatically determining whether two manuscript images were produced by the same scribe.  To this end, we explore a series of neural architectures that learn discriminative embeddings of handwriting images, enabling the measurement of similarity between fragments using distance metrics.
The experimental setup involves two datasets: the Tsinghua Bamboo Slips Dataset\cite{tsinghua2025}, which consists of digitized images of ancient Chinese manuscripts, and a curated subset of the Multi-Attribute Chinese Calligraphy Dataset (MCCD)\cite{mccd2025}, which contains samples from different calligraphers and provides abundant image data. These datasets offer a diverse range of handwriting styles, providing a foundation for training and evaluating models.

Several neural architectures have been implemented and compared, including Siamese networks trained with contrastive loss, Triplet networks optimized with triplet loss~\cite{triplet}, and ViT-based models~\cite{vit2020} for capturing both local and global handwriting features. 
Modern CNN backbones such as Visual Geometry Group (VGG19)~\cite{vgg19}, Residual Network (ResNet)~\cite{resnet}, and MobileNetV3~\cite{mobilenetv3}, as well as recent ViT~\cite{vit2020}, have been applied to handwriting analysis.

Overall, the paper establishes a unified PyTorch-based training and evaluation framework for scribe verification that can handle both ancient and modern datasets. It adopts a dynamic pair- and triplet-sampling mechanism, implements consistent preprocessing and normalization pipelines, and standardizes evaluation using fixed-pair generation and Receiver Operating Characteristic (ROC)-based analysis.
The paper's main contribution is a unified comparison of Siamese, Triplet, and ViT models for scribe verification across Chinese handwriting datasets. 
Unlike prior studies that typically focus on a single architecture or dataset, this work presents a unified and reproducible benchmarking framework for scribe verification across both ancient and modern Chinese handwriting datasets. While the individual model components and loss functions are well established, their systematic evaluation under identical preprocessing, dynamic sampling, and ROC-based evaluation protocols provides practical insights into architectural trade-offs for metric learning-based scribe verification. 

The remainder of this paper is organized as follows. Section II describes the datasets, model architectures, and training methodology. Section III presents the experimental results and comparative evaluation. Section IV concludes the paper and outlines directions for future research.

\section{Methodology} \label{sec:methodology}
\subsection{Datasets} \label{subsec:datasetDescription}

\begin{figure}[!t]
    \centering
    \begin{subfigure}{0.10\textwidth}
        \centering
        \includegraphics[width=\linewidth]{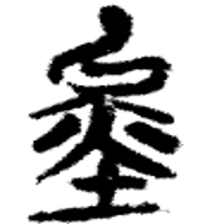}
        \caption{Bao Xun}
    \end{subfigure}
    \hspace{0.2cm}
    \begin{subfigure}{0.10\textwidth}
        \centering
        \includegraphics[width=\linewidth]{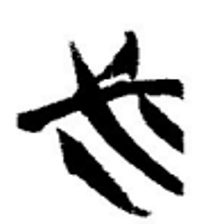}
        \caption{Guan Zhong}
    \end{subfigure}
    \caption{Representative examples from the Tsinghua Bamboo Slips dataset.}
    \label{fig:tsinghua_samples}

    \vspace{-0.4cm} 
\end{figure}

The Tsinghua Bamboo Slips dataset~\cite{tsinghua2025} consists of digitized images of ancient Chinese manuscripts discovered in 2008 at Tsinghua University. 
Bamboo slips were an early writing medium in ancient China, preceding the widespread adoption of paper. Text was written vertically on narrow bamboo strips using a brush and ink, and manuscripts were formed by binding multiple slips together. Individual scribes exhibit characteristic stroke patterns, spacing, and ink usage, making scribe verification a valuable task for manuscript reconstruction, authentication, and historical analysis.
Each fragment corresponds to a section of bamboo slip text written in classical Chinese, attributed to specific scribes. 
Two such examples are depicted in Figure~\ref{fig:tsinghua_samples}. The dataset serves as a historically significant benchmark for studying scribe handwriting patterns and presents challenges such as faded ink, uneven texture, and varying stroke thickness. These characteristics make it an ideal testbed for evaluating neural networks' ability to extract robust visual features from degraded manuscript data. The distribution of images per scribe is listed in Table~\ref{tab:tsinghua_per_scribe}. 

\begin{table}[!htb]
\centering
\caption{DISTRIBUTION OF IMAGES PER SCRIBE IN THE TSINGHUA BAMBOO SLIPS DATASET.}
\label{tab:tsinghua_per_scribe}
\begin{tabular}{lcc}
\toprule
\textbf{Scribe Class} & \textbf{Training Images} & \textbf{Test Images} \\
\midrule
Bao Xun & 164 & 40 \\
Bie Gua & 32 & 8 \\
Cheng Wu & 214 & 53 \\
Chu Ju & 445 & 111 \\
Feng Xu Zhi Ming & 161 & 40 \\
Guan Zhong & 1508 & 377 \\
Huang Men & 4282 & 1071 \\
Liang Chen & 285 & 71 \\
Ming Xun & 422 & 107 \\
Shi Fa & 1670 & 420 \\
Yin Zhi & 3412 & 852 \\
\midrule
\textbf{Total} & \textbf{12595} & \textbf{3050} \\
\bottomrule
\end{tabular}
\end{table}

\begin{figure}[!htb]
    \centering

    \begin{subfigure}{0.10\textwidth}
        \centering
        \includegraphics[width=\linewidth]{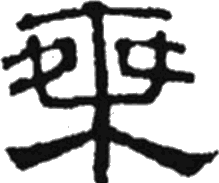}
        \caption{Wu Rui}
    \end{subfigure}
    \hspace{0.2cm}
    \begin{subfigure}{0.10\textwidth}
        \centering
        \includegraphics[width=\linewidth]{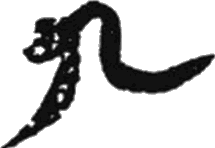}
        \caption{Huang Tingjian}
    \end{subfigure}

    \caption{Representative examples from the MCCD dataset.}
    \label{fig:MCCD_samples}

    \vspace{-0.4cm} 
\end{figure}

The MCCD~\cite{mccd2025} is a large-scale modern collection of handwritten Chinese character images created by multiple professional calligraphers.
Two representative examples are shown in Figure~\ref{fig:MCCD_samples}.
To adapt this dataset for the scribe verification task, a curated subset was created by selecting only the Calligraphers folder and including the scribes with the largest number of samples. 
This selection ensures a more balanced, representative distribution of handwriting styles, thereby avoiding extreme class imbalance. Each subfolder corresponds to a single calligrapher, and the images were used to form positive and negative training examples for Siamese and Triplet learning.
The distribution of images per scribe is listed in Table~\ref{tab:MCCD_per_scribe}.

\begin{table}[!htb]
\centering
\caption{DISTRIBUTION OF IMAGES PER SCRIBE IN THE MCCD SUBSET DATASET.}
\label{tab:MCCD_per_scribe}
\begin{tabular}{lcc}
\toprule
\textbf{Scribe Class} & \textbf{Training Images} & \textbf{Test Images} \\
\midrule
He Shaoji & 1048 & 448 \\
Shan Xiaotian & 1114 & 447 \\
Wu Rui & 991 & 424 \\
Wu Dacheng & 1242 & 532 \\
Mi Fu & 1526 & 654 \\
Deng Shiru & 923 & 395 \\
Yan Zhenqing & 1655 & 708 \\
Huang Tingjian & 1353 & 579 \\
\midrule
\textbf{Total} & \textbf{9852} & \textbf{4187} \\
\bottomrule
\end{tabular}
\end{table}

\subsection{Model Architectures}\label{subsec:modelArchitecture}

Each model is designed to learn an embedding representation of handwriting fragments such that samples from the same scribe are mapped close together in the embedding space. In contrast, samples from different scribes are widely separated. 
Three main architectural families are explored: convolutional Siamese networks, a Triplet network variant, and a ViT Siamese model.

\subsubsection{Siamese CNN-based Models}\label{subsubsec:siamese}
The first set of experiments employed convolutional Siamese architectures trained with the contrastive loss function. A Siamese network consists of two identical branches that share weights and process two input images in parallel. Each branch extracts features from an input image and outputs a low-dimensional embedding vector. The distance between these embeddings, typically measured using the Euclidean distance, is minimized for samples belonging to the same scribe and maximized for samples from different scribes.

Several pretrained convolutional backbones were used as feature extractors, including MobileNetV3 \cite{mobilenetv3}, MobileNetV3+ Custom \cite{tsinghua2025}, VGG19 \cite{vgg19}, and ResNet18/34/50 \cite{resnet}. The final layers of these networks were modified to produce a 10-dimensional embedding vector, consistent with the dimensionality used in the paper on scribe verification~\cite{tsinghua2025}. These embeddings were learned by fine-tuning pretrained ImageNet weights, enabling the models to adapt to the specific texture and style features of handwritten Chinese characters. 

Among the CNN-based Siamese models, the MobileNetV3+ Custom architecture achieved the best performance on the Tsinghua Bamboo Slips dataset when trained with the contrastive loss. Its lightweight structure, combined with efficient depthwise separable convolutions, allowed it to learn discriminative handwriting representations without overfitting, even with a limited number of samples per scribe. 

\subsubsection{Triplet Network}\label{subsubsec:triplet}
The second model was based on the Triplet learning paradigm~\cite{triplet}, 
a metric-learning approach designed to learn embeddings in which examples from the same class lie closer together than examples from different classes. Unlike Siamese networks, which operate on image pairs, triplet networks process three inputs simultaneously: an anchor sample, a positive sample (from the same scribe as the anchor), and a negative sample (from a different scribe). 

Triplet learning is beneficial for problems where the goal is not classification but verification or similarity comparison, such as writer identification.   By explicitly modeling relative similarity rather than absolute labels, the network learns fine-grained variations within each scribe's writing style while maximizing separation between different scribes.

The Triplet architecture used in the paper was derived from the MobileNetV3+ Custom backbone. The shared embedding network outputs a 10-dimensional feature vector for each input, and the loss is computed according to the triplet constraint. Although this approach improves the separation among different scribe classes, it requires significantly more computational time and careful sampling of informative triplets.

\subsubsection{ViT Siamese Model}\label{subsubsec:vit}
To investigate the effectiveness of self-attention mechanisms in scribe verification, a ViT ~\cite{vit2020} was implemented within the Siamese framework. Unlike convolutional networks, which operate on local receptive fields, ViT processes images as sequences of non-overlapping patches and captures global contextual relationships through multi-head self-attention layers. This property enables the model to understand better the overall structure and spatial relationships in handwritten text.

The ViT-B/16 architecture, pretrained on ImageNet, served as the backbone, with its final classification head replaced by a fully connected layer projecting to a 10-dimensional embedding space. Both branches of the ViT Siamese model share weights, and training is performed using the same contrastive loss formulation as in the CNN-based Siamese models. 

\subsection{Loss Functions}\label{subsec:lossFunctions}
Deep metric learning aims to project input images into an embedding space where distances correspond to semantic similarity. Two main loss functions were tested to guide the learning process: the Contrastive Loss and the Triplet Loss. 
Both are designed to minimize the distance between embeddings of samples from the same scribe while maximizing the distance between samples from different scribes.

\subsubsection{Contrastive Loss}\label{subsubsec:contrastiveLoss}
The contrastive loss function~\cite{tsinghua2025} was used to train the Siamese and Vision Transformer models. Given a pair of images $(x_1, x_2)$ with label $y \in \{0,1\}$, where $y=1$ indicates that both images belong to the same scribe and $y=0$ indicates different scribes, the Euclidean distance between their embeddings is defined as  $D(x_1,x_2) = \|f(x_1) - f(x_2)\|_2$. 
The loss is given by:
\begin{align} 
L_{contrastive}(x_1,x_2) &= \frac{1}{2} \, y \, D^{2}(x_1,x_2) + \frac{1}{2} (1 - y) \cdot \nonumber \\ & \;\; \cdot \max\left(0, m - D(x_1,x_2) \right)^{2}, \label{Eq:contrastive}
\end{align}
where $m$ is a predefined margin that controls the separation between positive and negative pairs. 
The first term in (\ref{Eq:contrastive}) penalizes large distances for positive pairs (same scribe), while the second term penalizes small distances for negative pairs that lie within the margin. 
This formulation encourages embeddings of the same scribe to cluster tightly together while maintaining a minimum distance between different scribes.

\subsubsection{Triplet Loss}\label{subsubsec:tripletLoss}
The triplet loss was applied to the Triplet Network~\cite{triplet}, which processes three images simultaneously: an anchor $x_a$, a positive $x_p$ (same scribe), and a negative $x_n$ (different scribe). The goal is to enforce that the distance between the anchor and the positive sample is smaller than the distance between the anchor and the negative sample by at least a margin $m$. 
Formally, the loss is expressed as:
\begin{equation}\label{Eq:triplet}
L_{triplet}(x_a, x_p, x_n) = \max \big( 0, D(x_a,x_p) - D(x_a,x_n) + m \big),
\end{equation}
where $D(x_a,x_p)$ and $D(x_a,x_n)$ denote the Euclidean distances between the corresponding embeddings. 
The objective (\ref{Eq:triplet}) helps the model to create a structured embedding space in which clusters of samples belonging to the same scribe are well separated from those of different scribes.

\subsection{Data Preprocessing and Sampling Strategy}\label{subsubsec:dataPreprocessing}
All manuscript images underwent standardized preprocessing and sampling to ensure consistency across datasets and to maximize intra-class variability during training.
Each image was first converted to a three-channel RGB format, regardless of its original grayscale input, to maintain compatibility with pretrained ImageNet backbones. 
All images were then resized to $224\times224$ pixels, following the configuration used in the Tsinghua Bamboo Slip experiments in~\cite{tsinghua2025}. 
Normalization was applied using the ImageNet mean $\mu = (0.485, 0.456, 0.406)$ and standard deviation $\sigma = (0.229, 0.224, 0.225)$.
These preprocessing steps were implemented within the \texttt{PyTorch} transformation pipeline to ensure consistent input scaling across all architectures.

To improve generalization, several lightweight augmentations were applied randomly during training: 1) Horizontal flipping to simulate mirrored character structures; Random grayscale inversion (gray-flip) for minority classes, following the data balancing strategy described in the Tsinghua framework; 3) Random brightness and contrast adjustments within a small range.
This combination increased variability in stroke density and ink appearance, which is important for robust learning of scribe representations.

Training samples were generated dynamically at runtime, such as Positive pairs: two images from the same scribe or calligrapher folder; Negative pairs: two images from different scribes; and Triplets: an anchor, a positive from the same scribe, and a negative from a different scribe.
Positive and negative pairs were balanced using a $1\!:\!1$ ratio to prevent bias toward a particular class type. 
For triplet training, each batch contained randomly sampled anchor–positive–negative combinations generated on the fly. 
This dynamic sampling approach eliminates the need for storing predefined pairs and increases the diversity of training examples.

During dataset loading, corrupted or unreadable image files were automatically detected and replaced with a resampled, valid image from the same class. If repeated failures occurred, a blank placeholder image was substituted to maintain batch consistency and prevent runtime interruptions. 
In practice, image loading failures were rare (occurring in fewer than 0.5\% of samples) and were primarily caused by corrupted files. The placeholder mechanism was therefore included as a safeguard to prevent training interruptions rather than as a frequently used operation.

For evaluation, fixed image pairs were generated using a separate script, yielding balanced sets of positive and negative examples. Each pair was labeled as ``same'' or ``different'' depending on whether both images originated from the same scribe. This ensured reproducible and unbiased comparisons across different architectures and training configurations.

\subsection{Training Setup}\label{subsec:trainingSetup}
All models were implemented and trained using the \texttt{PyTorch} deep learning framework. 
The experiments were conducted in a controlled environment to ensure reproducibility and consistency across architectures and datasets.

Training was performed exclusively on an NVIDIA GeForce RTX 4060 Laptop GPU with 8 GB of VRAM, supported by an Intel Core i7-13620H processor and 32 GB of system memory. 
All computations were executed on the GPU using CUDA acceleration, significantly reducing training time and enabling parallelized batch processing for both Siamese and Triplet configurations.

The experiments were executed within a dedicated \texttt{Python} virtual environment (\texttt{.venvsiamese}) using \texttt{Python 3.10} and \texttt{PyTorch 2.1.0} with \texttt{TorchVision 0.16.0}. 
Supporting libraries included \texttt{NumPy}, \texttt{Pillow}, \texttt{tqdm}, and \texttt{PyYAML} for data handling, progress monitoring, and configuration management. 
All training parameters were stored in \texttt{YAML} configuration files, which defined dataset paths, model settings, and optimizer hyperparameters.

All models were trained using the Adam optimizer with an initial learning rate of $1\times10^{-3}$ and no weight decay. 
A batch size of~32 was employed, and each training session ran for 30~epochs. 
A constant learning rate was maintained throughout, as this setting yielded stable convergence across architectures without requiring a scheduler.

The margin hyperparameter for the contrastive loss was fixed at $m = 0.6$, following the configuration of the Tsinghua Bamboo Slip framework. 
For the triplet loss, the margin was set to $m = 1.0$. 
All models produced a 10-dimensional embedding vector at the output layer, consistent with the reference design.

At each iteration, positive and negative pairs (or triplets) were generated dynamically to ensure exposure to new sample combinations. 
The Euclidean distance between embeddings was computed, and the appropriate loss function—contrastive or triplet—was applied. 
Model weights were updated via backpropagation, and checkpoints were saved after every epoch in the \texttt{checkpoints\ directory} using the format \texttt{model\_e\{epoch\}.pt}. 
CUDA computation enabled efficient mini-batch parallelization and faster convergence across all model variants.

A fixed random seed ($42$) was applied globally to dataset loading, pair sampling, and model initialization to guarantee reproducibility of training outcomes. 

While the main evaluation was conducted after full training, intermediate validation losses were monitored to verify stable convergence and to detect potential overfitting. 
The best-performing model checkpoint, based on validation stability, was selected for the final evaluation on the fixed test-pair set.
The code implementing the scribe verification methods is available in \cite{Scribe_code}.
Neither the Tsinghua Bamboo Slips Dataset nor the MCCD dataset is publicly available. Still, they can be accessed upon reasonable request for research purposes~\cite{tsinghua2025,mccd2025}. 
Only publicly released or author-approved samples were used in the paper, and no restricted or unpublished manuscript materials were distributed.

\subsection{Evaluation Procedure}\label{subsec:evaluationProcedure}
After training, each model was evaluated on a fixed test set composed of predefined image pairs. 
These pairs were generated using a dedicated Python script (\texttt{make\_test\_pairs.py}) that created a balanced dataset of positive (same scribe) and negative (different scribe) examples. 
The resulting list was stored in a CSV file named \texttt{test\_pairs.csv}, containing three columns: the file paths of the two images (\texttt{path1} and \texttt{path2}) and the binary label (\texttt{label}), where 1~denotes a same-scribe pair and 0~denotes a different-scribe pair. The \texttt{CSV} file ensures that all models are evaluated on the same test samples and labels, thereby enabling reproducible, unbiased comparisons across architectures.

During evaluation, each input image was passed through the trained network to obtain its corresponding embedding vector $f(x) \in \mathbb{R}^{10}$. 
For Siamese and Vision Transformer models, embeddings were generated for both images in each pair, and the similarity between them was computed using the Euclidean distance.
Smaller distances indicate higher similarity between the two fragments, suggesting that the same scribe likely wrote them.

A decision threshold $\tau$ was applied to classify whether two fragments $x_1$ and $x_2$ were written by the same or different scribes:
\begin{equation}
\text{prediction} =
\begin{cases}
\text{same scribe}, & D(x_1,x_2) < \tau, \\
\text{different scribes}, & D(x_1, x_2) \ge \tau.
\end{cases}
\end{equation}
The optimal threshold was determined by scanning possible values and selecting the one that maximized overall classification accuracy. This approach aligns with the evaluation methodology presented in the Tsinghua Bamboo Slip framework \cite{tsinghua2025}.

To assess the discriminative power of the learned embeddings, several quantitative metrics were computed:
1) \textit{ROC curve}, showing the trade-off between true and false acceptance rates. In the context of scribe verification, the ROC curve provides insight into how well the model separates same-scribe pairs from different-scribe pairs across all possible decision thresholds. 2) \textit{
Area Under the ROC Curve (AUC)}, summarizing ROC performance by a single scalar value. The AUC offers a threshold-independent measure of discriminative ability. 3) \textit{Accuracy (ACC)}, representing the proportion of correctly classified pairs. 4) \textit{False Acceptance Rate (FAR)}, quantifying the ratio of negative pairs incorrectly predicted as positive ones (i.e., same). 5) \textit{False Rejection Rate (FRR) or False Negative Rate}, quantifying the ratio of positive pairs incorrectly predicted as negative (i.e., different).
These metrics were computed using \texttt{NumPy} and \texttt{scikit-learn} functions implemented in the evaluation pipeline. 
The ROC curves were plotted using \texttt{Matplotlib} and saved as visual summaries for model comparison.

The evaluation script (\texttt{evaluate.py} or \texttt{evaluate\_triplet.py}) automatically loads the trained checkpoint from \texttt{checkpoints\ directory}, reads the \texttt{test\_pairs.csv} file, extracts embeddings for each pair, computes the Euclidean distances, and evaluates the metrics described above. 
The results are printed to the console and visualized as ROC curves, while detailed statistics, including AUC, accuracy, and the optimal decision threshold, are recorded. 
This standardized evaluation procedure ensures that all models are compared fairly and under identical test conditions.

\section{Experiments and Results} \label{sec:experimentsAndResults}

Experiments were conducted on both the Tsinghua Bamboo Slips dataset and the MCCD subset, following the methodology described in Section~\ref{sec:methodology}. 

Table~\ref{tab:results_all} summarizes the quantitative performance of all trained architectures on the Tsinghua and MCCD datasets. 
The MobileNetV3+~Custom Siamese model trained with contrastive loss achieved the best overall results, obtaining an AUC of~0.958 and an accuracy of 89.19\% in the Tsinghua Bamboo Slips Dataset. In contrast, the ResNet34 model trained with contrastive loss achieved the best performance on the MCCD Dataset, achieving an AUC of ~0.952 and an accuracy of 89.06\%.
Unlike the standard MobileNetV3, which ends with a large classification head, the MobileNetV3 Custom version replaces this component with a reduced 160-channel bottleneck and a final 10-dimensional embedding layer, making it better suited for distance-based metric learning in scribe verification.

\begin{table}[!t]
\centering
\scriptsize
\setlength{\tabcolsep}{2.2pt}
\renewcommand{\arraystretch}{0.9}
\caption{QUANTITATIVE PERFORMANCE COMPARISON OF ALL TRAINED ARCHITECTURES ON THE EVALUATED DATASETS.}
\label{tab:results_all}
\begin{tabular}{lccccc}
\toprule
\textbf{Model} & \textbf{Dataset} & \textbf{AUC} & \textbf{ACC} (\%)& \textbf{FAR} (\%) & \textbf{FRR} (\%)\\
\midrule
MobileNetV3 (Siamese) & Tsinghua & 0.897 & 82.27 & 13.28  & 22.19 \\
\textbf{MobileNetV3+ Custom (Siamese)} & Tsinghua & \textbf{0.958} & \textbf{89.19} & \textbf{5.63} & \textbf{15.99} \\
ResNet18 (Siamese) & Tsinghua & 0.921 & 84.65 & 16.56 & 14.15 \\
ResNet34 (Siamese) & Tsinghua & 0.912 & 82.56 & 19.5 & 15.37 \\
ResNet50 (Siamese) & Tsinghua & 0.884 & 80.4 & 21.5 & 17.69 \\
VGG19 (Siamese) & Tsinghua & 0.857  & 82.07 & 14.95 & 20.91 \\
ViT-B/16 (Siamese) & Tsinghua & 0.829  & 77 & 25.59 & 20.41 \\
MobileNetV3+ Custom (Triplet) & Tsinghua & 0.945 & 88.16 & 13.59  & 10.15 \\
\midrule
MobileNetV3 (Siamese) & MCCD & 0.949 & 88.99  & 11.44  & 10.59 \\
MobileNetV3+ Custom (Siamese) & MCCD & 0.949 & 88.5 & 12.77 & 10.23 \\
ResNet18 (Siamese) & MCCD & 0.948 & 88.24 & 13.93 & 9.58 \\
\textbf{ResNet34 (Siamese)} & \textbf{MCCD} & \textbf{0.952} & \textbf{89.06} & \textbf{14.08} & \textbf{7.79} \\
ResNet50 (Siamese) & MCCD & 0.931 & 86.93 & 17.91 & 8.23 \\
VGG19 (Siamese) & MCCD & 0.9  & 82.93 & 25.05 & 9.10 \\
ViT-B/16 (Siamese) & MCCD & 0.89  & 82.06 & 24.41 & 11.47 \\
MobileNetV3+ Custom (Triplet) & MCCD & 0.931 & 87.63 & 16.49  & 8.66 \\
\bottomrule
\end{tabular}
\end{table}

\begin{figure}[!htb]
\centering
\includegraphics[width=0.5\linewidth]{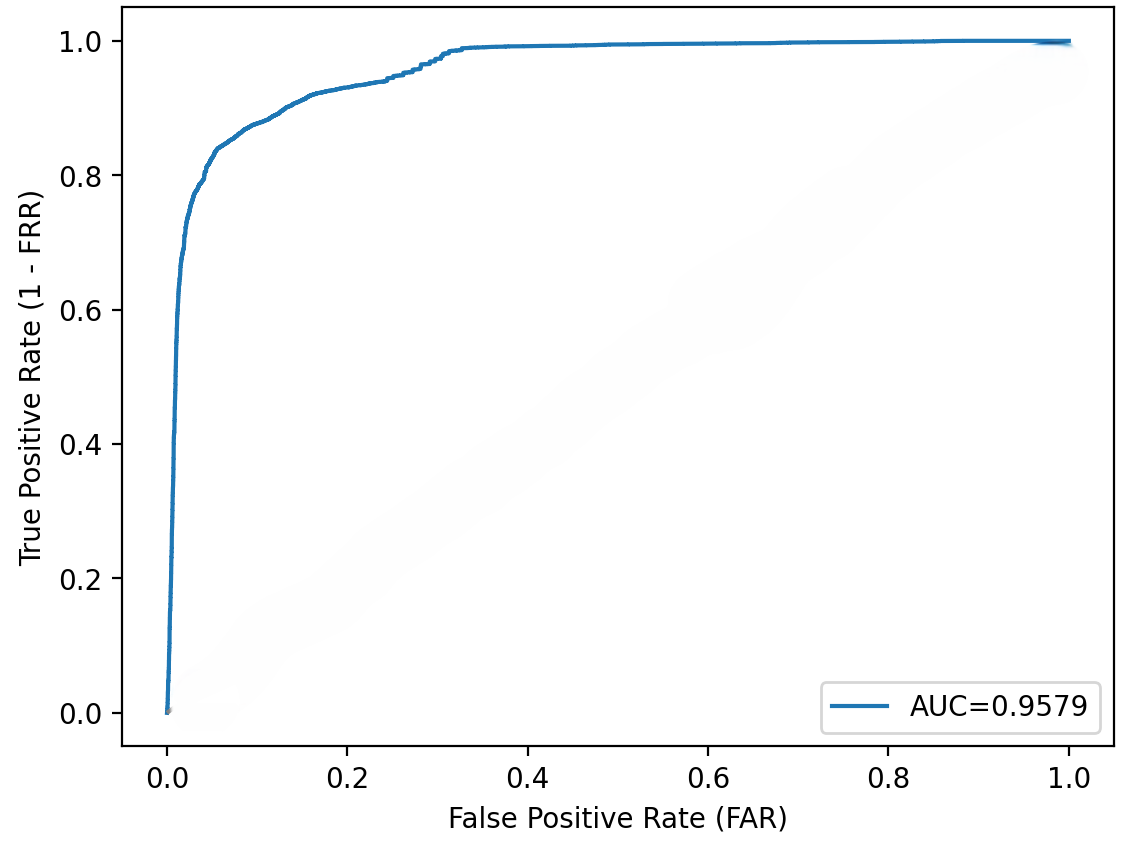}
\caption{ROC curve of the MobileNetV3+ Custom Siamese model on the Tsinghua Bamboo Slips dataset.}
\label{fig:roc_mobilenet_tsinghua}
\end{figure}

\begin{figure}[!htb]
\centering
\includegraphics[width=0.5\linewidth]{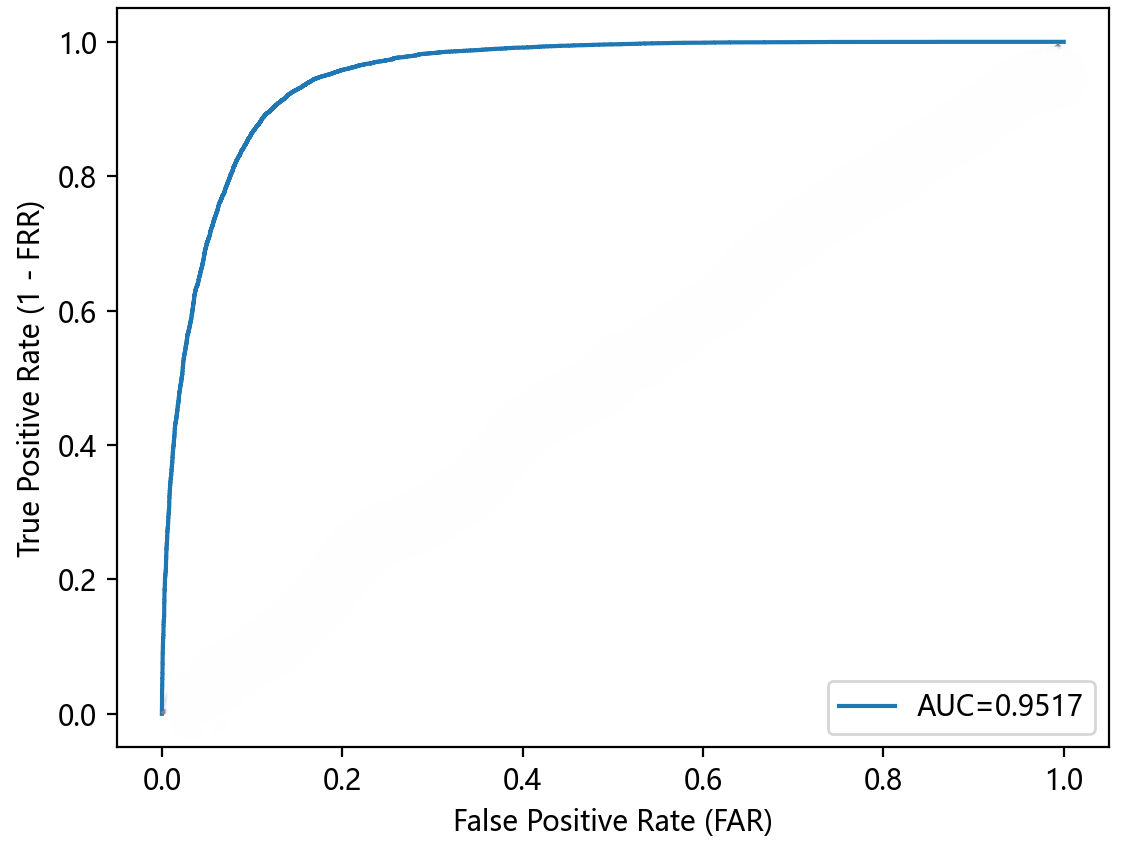}
\caption{ROC curve of the ResNet34 Custom Siamese model on the MCCD dataset.}
\label{fig:roc_ResNet34_MCCD}
\end{figure}

Figures~\ref{fig:roc_mobilenet_tsinghua} and \ref{fig:roc_ResNet34_MCCD}  
depict the ROC curves of representative models. 

The experimental results demonstrate that deep metric approaches are effective for scribe verification across both ancient and modern Chinese handwriting datasets. 
However, the optimal network configuration varies depending on the characteristics and quality of the dataset. Overall, convolutional Siamese architectures produced the most stable and discriminative embeddings, while Triplet and Transformer-based variants provided additional insights but did not surpass the baseline models.

On the Tsinghua dataset, the MobileNetV3+ Custom achieved the highest overall performance among all tested architectures, as shown in Table~\ref{tab:results_all}. 
This success can be attributed to its ability to balance depth and generalization, effectively
capturing mid-level texture features crucial for distinguishing writing patterns on aged bamboo fragments. That is, MobileNetV3+ Custom exhibits a strong inductive bias toward local texture and stroke-level features, which are critical in degraded manuscript images.
The model’s moderate complexity enabled it to handle noise, surface irregularities, and faded ink without overfitting, a challenge observed in deeper networks such as ResNet-50 and VGG-19. 

Medium-depth convolutional architectures balance representational capacity and generalization, enabling them to capture discriminative handwriting patterns without overfitting to noise. In contrast, deeper CNNs and transformer-based models may be more sensitive to background artifacts and require substantially larger datasets to learn robust representations.

On the modern MCCD dataset, the ResNet34 Siamese model achieved the best results, achieving the highest AUC and accuracy among all tested configurations. The ResNet 34 Siamese model also performed exceptionally well on the Tsinghua Bamboo Slips dataset. Its structure and the efficient feature extraction were advantageous for the cleaner, high-resolution images of modern calligraphers. 

The strong performance of MobileNetV3+ on the MCCD dataset suggests that efficient CNNs may be sufficient for capturing stylistic differences in consistent, well-preserved handwriting.  
Future work will include evaluating the ResNet18 backbone on the MCCD dataset to validate this hypothesis.

The Triplet Network based on the MobileNetV3+ Custom backbone did not outperform its Siamese counterpart. 
Despite theoretically promoting a more structured embedding space, the triplet configuration required considerably longer training and yielded slightly lower accuracy and AUC.
This indicates that, for the datasets and sampling strategies used in this study, the contrastive loss function yielded more stable optimization and better generalization.

Direct comparison with previously published results is challenging due to differences in dataset splits, pair generation strategies, and the limited public availability of the evaluated datasets. Nevertheless, the observed performance trends are consistent with prior findings reported in \cite{tsinghua2025}, which showed that medium-depth convolutional Siamese networks outperform deeper CNN architectures on degraded ancient manuscript images.

\section{Conclusion and Future Work}

The contrastive Siamese networks have been highly effective for this task, achieving robust performance across both datasets. 
The MobileNetV3+ Custom Siamese model has achieved the top performance on the Tsinghua Bamboo Slips dataset despite challenging image degradation. 
On the modern MCCD dataset, the ResNet34 Siamese model also achieved the highest overall accuracy and AUC, confirming the suitability of lightweight architectures for high-quality calligraphy data. 

Future research will focus on several directions: 1) Exploring hybrid CNN–Transformer architectures that combine efficient local feature extraction with global attention mechanisms; 2) Incorporating hard positive and hard negative mining strategies to improve triplet-based training efficiency; 3) Expanding the dataset with additional manuscript sources to enable large-scale pretraining and improve generalization;  4) Investigating few-shot and one-shot learning approaches to scribe verification with minimal labeled data; and
5) Including multi-seed experiments to quantify performance variability and further assess the robustness of metric learning methods that are sensitive to initialization and sampling strategies.

\section*{Acknowledgment}
We want to thank Professor Lianwen Jin from the School of Electronic and Information Engineering at South China University of Technology, Guangzhou, China, for sharing the MCCD dataset with us.

\bibliographystyle{IEEEtran}
\bibliography{References}

\begin{thebibliography}{10}
\providecommand{\url}[1]{#1}
\csname url@samestyle\endcsname
\providecommand{\newblock}{\relax}
\providecommand{\bibinfo}[2]{#2}
\providecommand{\BIBentrySTDinterwordspacing}{\spaceskip=0pt\relax}
\providecommand{\BIBentryALTinterwordstretchfactor}{4}
\providecommand{\BIBentryALTinterwordspacing}{\spaceskip=\fontdimen2\font plus
\BIBentryALTinterwordstretchfactor\fontdimen3\font minus \fontdimen4\font\relax}
\providecommand{\BIBforeignlanguage}[2]{{%
\expandafter\ifx\csname l@#1\endcsname\relax
\typeout{** WARNING: IEEEtran.bst: No hyphenation pattern has been}%
\typeout{** loaded for the language `#1'. Using the pattern for}%
\typeout{** the default language instead.}%
\else
\language=\csname l@#1\endcsname
\fi
#2}}
\providecommand{\BIBdecl}{\relax}
\BIBdecl

\bibitem{Diao2025}
\BIBentryALTinterwordspacing
D.~Xiaolei \emph{et~al.}, ``Ancient script image recognition and processing: A review,'' 2025, [retrieved: January 31, 2026]. [Online]. Available: \url{https://arxiv.org/abs/2506.19208}
\BIBentrySTDinterwordspacing

\bibitem{Qin2017}
H.~Qin and L.~Peng, ``Convolutional neural network with attention mechanism for historical chinese character recognition,'' in \emph{Proceedings of the 4th International Workshop on Historical Document Imaging and Processing}.\hskip 1em plus 0.5em minus 0.4em\relax Kyoto, Japan: {ACM}, 2017, pp. 42--47.

\bibitem{Markou2021}
K.~Markou \emph{et~al.}, ``A convolutional recurrent neural network for the handwritten text recognition of historical greek manuscripts,'' in \emph{Pattern Recognition. ICPR International Workshops and Challenges: Virtual Event, January 10-15, 2021, Proceedings, Part VII}.\hskip 1em plus 0.5em minus 0.4em\relax Berlin, Heidelberg: Springer-Verlag, 2021, pp. 249--–262.

\bibitem{Wu2021}
L.~Wu, C.~Zhang, M.~Xu, and M.~Wu, ``Ancient chinese recognition method based on attention mechanism,'' in \emph{Proceedings of the 7th IEEE International Conference on Network Intelligence and Digital Content (IC-NIDC)}, Beijing, China, 2021, pp. 309--313.

\bibitem{Sivan2025}
R.~Sivan, P.~B. Pati, and M.~W.~A. Kesiman, ``Image quality determination of palm leaf heritage documents using integrated discrete cosine transform features with vision transformer,'' \emph{International Journal on Document Analysis and Recognition}, vol.~28, no.~1, pp. 41--–57, July 2025.

\bibitem{tsinghua2025}
H.~Wang \emph{et~al.}, ``Tsinghua bamboo slip scribe verification using {Siamese} networks,'' \emph{Heritage Science}, 2025.

\bibitem{mccd2025}
\BIBentryALTinterwordspacing
Y.~Zhao, Y.~Zhang, and L.~Jin, ``{MCCD}: A multi-attribute {Chinese} calligraphy character dataset annotated with script styles, dynasties, and calligraphers,'' 2025, [retrieved: January 31, 2026]. [Online]. Available: \url{https://arxiv.org/abs/2507.06948}
\BIBentrySTDinterwordspacing

\bibitem{triplet}
\BIBentryALTinterwordspacing
E.~Hoffer and N.~Ailon, ``Deep metric learning using triplet network,'' 2014, [retrieved: January 31, 2026]. [Online]. Available: \url{https://arxiv.org/abs/1412.6622}
\BIBentrySTDinterwordspacing

\bibitem{vit2020}
\BIBentryALTinterwordspacing
A.~Dosovitskiy \emph{et~al.}, ``An image is worth 16$\times$16 words: Transformers for image recognition at scale,'' 2020, [retrieved: January 31, 2026]. [Online]. Available: \url{https://arxiv.org/abs/2010.11929}
\BIBentrySTDinterwordspacing

\bibitem{vgg19}
\BIBentryALTinterwordspacing
K.~Simonyan and A.~Zisserman, ``Very deep convolutional networks for large-scale image recognition,'' 2014, [retrieved: January 31, 2026]. [Online]. Available: \url{https://arxiv.org/abs/1409.1556}
\BIBentrySTDinterwordspacing

\bibitem{resnet}
K.~He, X.~Zhang, S.~Ren, and J.~Sun, ``Deep residual learning for image recognition,'' in \emph{Proceedings of the IEEE Conference on Computer Vision and Pattern Recognition}, Las Vegas, NV, USA, 2016, pp. 770--778.

\bibitem{mobilenetv3}
A.~Howard \emph{et~al.}, ``Searching for {MobileNetV3},'' in \emph{Proceedings of the IEEE/CVF International Conference on Computer Vision}, Seoul, South Korea, 2019, pp. 1314--1324.

\bibitem{Scribe_code}
\BIBentryALTinterwordspacing
``{Scribe Verification Using Siamese, Triplet, and Vision Transformer Networks},'' [retrieved: January 30, 2025]. [Online]. Available: \url{https://github.com/dimiliakop/ScribeVerification.git}
\BIBentrySTDinterwordspacing

\end{thebibliography}

\end{document}